\documentclass[12pt]{article}
\usepackage{mathpazo}
\usepackage{amsmath}
\usepackage{amssymb}
\usepackage{amsthm}
\usepackage{mathtools}
\usepackage{thmtools}
\usepackage{enumitem}
\usepackage{caption}
\usepackage{subcaption}
\usepackage{xcolor}
\usepackage{soul, accents}
\usepackage{tikz}
\usepackage{booktabs}
\usepackage{multirow}
\usepackage{makecell}
\usepackage{setspace}
\usepackage[labelsep=period,font=small,labelfont=bf,margin=5ex]{caption}
\usepackage[margin=1in]{geometry}
\usepackage[colorlinks=true,linkcolor=black,citecolor=black]{hyperref}
\usepackage{cleveref}
\usepackage[citestyle=authoryear,natbib=true,backend=biber]{biblatex}
\declaretheoremstyle[
headfont=\sffamily\bfseries,
bodyfont=\itshape,
spaceabove=12pt,
spacebelow=12pt
]{bktheorem}

\title{Diagnosing Hate Speech Classification: Where Do Humans and Machines Disagree, and Why?}

\author{Xilin Yang\footnote{Senior Research Specialist at Princeton University: \href{mailto:xy4772@princeton.edu}{xy4772@princeton.edu}}}
\date{\today}

\begin{document}

\maketitle


\textcolor{red}{\textbf{Note: This paper contains hate speech instances that are racially offensive and may cause discomfort in reading, and for this we offer our sincere apologies. While they are unavoidable due to the nature of this study, we will do our best to eliminate the appearances of these instances in their original forms.}}

\begin{abstract}
This study uses the cosine similarity ratio, embedding regression, and manual re-annotation to diagnose hate speech classification. We use a dataset \textit{Measuring Hate Speech} that contains 135,556 annotated comments on social media. We begin with exploring the inconsistency of human annotation from the dataset. Using embedding regression as a basic diagnostic, we found that female annotators are more sensitive to racial slurs that target the black population. Examples like this illustrate that human evaluations of hate speech content vary based on their identity. We perform a more complicated diagnostic by training a hate speech classifier using the best-performing pre-trained large language model(LLM) NV-Embed-v2 on the MTEB benchmark (ranked No.1 as of Aug 30, 2024). We convert texts to embeddings and run a logistic regression, and our classifier achieves a testing accuracy of 94\%. Using manual re-annotation, we find that machines make fewer mistakes than humans despite the fact that human annotations are treated as ground truth in the training set. Pairing manual annotation and cosine similarity ratio, we find that machines perform better than humans in correctly labeling long statements but perform worse in labeling short and obvious instances of racial slurs. We hypothesize that Aristotle's virtue as a mean and lesser evil as good may play a role in restricting the latest models’ ability to detect obvious malicious content. Humans may, for their own moral restrictions and fear of risks, curate data or align the model as their means to choose the "lesser evil." This consideration is important for future studies on the risks of AI.

\end{abstract}

\newpage

\begin{spacing}{1.5}
\section{Introduction}
\label{sec:research-purpose}
Natural language processing works by converting words into vectors that encode their semantic meanings. Before digging into the rigorous method of this paper, we would like to make an audacious claim that we each has our own large language model (LLM), which is our mind, and it is in each of our mind that we encode words differently based on our identity, memory, and attachments. Hence, a certain word may arouse very different feelings across individuals. To understand the disagreement between humans and machines in classifying speech, we must at some point hit a granular level such that it is our "embeddings" that get compared. This is the motivation for our research, a ``word for word" study where human-machine disagreement takes place at the same granular level. Very few studies in social science and computer science have reached this level of subtlety in understanding where and why LLMs and humans disagree with each other.

This motivation affects our choice of research tools: we expect to use embedding regression, a method of word embeddings, to peak into our mind and spot the subtle differences in human annotators' evaluations of words. While there are voluminous psychological studies on how humans' response time differs when viewing different terms, that difference is not as direct as we wish for a concrete evidence. We choose embedding regression because it shows us not deductive but conclusive variance in human responses to a certain word based on their identities.

Once we obtain this variance in human annotations, then we want to explore the following questions: In a estimable sense, how much does this variance affect the model performance in classifying hate speech? In an intuitive sense, how much does it factor into the training of the model before we even begin execute a classification task? In other words, how do our identity, memory, and attachments, distinctive reactions to toxic terms factor into the birth of the LLMs? In a philosophical sense, how much do we want the LLMs to ``be like'' us, not only based on how similar they are to the way we speak, but also based on how they reflect our embedded moral restrictions before we speak?

Our study is an undoubted fruit of social science and humanities because many patterns, whether they are the variance in human annotations or the restrictions on the LLMs, shed lights on established social theory. We hope that by the end of this paper we can show you that the ethical challenges of AI are just an mirror image of our own struggle of humanity - a swinging between the conscious or unconscious choices of the good, the evil, and oftentimes, the lesser evil.


\section{Background}

\subsection{Word embeddings}
The distributional hypothesis~\citep{doi:10.1080/00437956.1954.11659520} from Linguistics dictates that words used in the same contexts tend to have similar meanings.
Based on this observation, word embedding models like GloVe~\citep{pennington-etal-2014-glove} and fastText~\citep{bojanowski2016enriching} map each word to a numerical vector, such that similar vectors represent words that appear in similar contexts, which further implies they have similar meanings.
In particular, we measure the similarity of two vectors $a$ and $b$ by cosine similarity: $\cos\left(\theta_{a, b}\right) = \frac{a \cdot b}{\lVert a \rVert \lVert b \rVert}$.

\subsection{ALC embeddings}
The same word in different text corpus may have different meanings.
For example, ``football'' means ``soccer'' in British English but ``American football'' in American English.
Suppose we are given a word embedding model trained on British English, how can we use it to understand American English?

\cite{arora-etal-2016-latent,arora-etal-2018-linear} show that under some plausible assumptions, one can obtain an embedding for any word in a new context by taking the average of some pre-trained embeddings of the words around it, and applying a linear transformation.
This is known as the ``a la carte'' embedding, or ALC embeddings.
In this paper, we use the ALC embeddings based on \texttt{fastText} for English prepared by \cite{204871}.

\subsection{Embedding regression}

Suppose we have a dataset containing a list of articles (denoted $t_i$) and whether each one is in British or American English (denoted $y_i$), and we want to know if there is any difference between the semantic meaning of a word ``football'' across these two dialects.
One way to achieve this goal is to do a multivariate regression with $y_i$ as the regressor, and the ALC embedding of $w$ computed on $t_i$ as the response.

More concretely, we have
\begin{equation}
\operatorname{ALC}_\text{football}(t_i) = \beta_0 + \beta_1 y_i + \epsilon_i,
\end{equation}
where $\operatorname{ALC}_\text{football}(\cdot)$ is the ALC embedding operator for ``football'', $\beta_0$ and $\beta_1$ are regression coefficients with the same dimension as the word embeddings, and $\epsilon_i$ stands for the noise vector. For example, if the meaning of ``football'' is similar across British or American English, then $y_i$ should not have a significant impact on $\operatorname{ALC}_\text{football}(t_i)$, which means $\beta_1$ should be close to zero.
Hence, we take the norm of $\beta_1$ and run a permutation test to see whether it's significant different from zero.
This is the idea of embedding regression~\citep{RODRIGUEZ_SPIRLING_STEWART_2023}.

Our use of embedding regression strictly follows the interpretation from~\cite{RODRIGUEZ_SPIRLING_STEWART_2023} in that ``we will make no claims that the distributions per se have causal effects on human understandings of terms. Thus, when we speak of the meaning of a focal word being different across groups, we are talking in a thin sense about the distribution of other words within a fixed window size of that focal word being different.''
This is important to keep in mind when we interpret our regression results: we make no claims that our cosine distances indicate any indication of words, but they do show a simple but informative comparison of language patterns.

\section{Data Collection}
\label{sec:data-collection}
The primary data set we study is \textit{Measuring Hate Speech}, which contains 135,556 annotated comments, which not only indicate whether a comment is hate speech, but also labels the type of hate speech (i.e., racially discriminatory, sexually offensive) as well as the annotators' demographic characteristics.

\section{Inconsistency in Human Annotation}

We devote this part of the paper to evaluating the nature of hand coding through embedding regression, by doing so we address the following concern: in the online environment, the use of many slurs inside a social group conveys drastically different meanings from the use outside that group. Will annotators' identity influence the result of annotation when labeling this kind of slurs?

\subsection{Embedding Regression}
We run embedding regression~\citep{RODRIGUEZ_SPIRLING_STEWART_2023} on a collection of 1,101,165 online comments which include the ones pre-classified as hate speech. To capture the content of hate, we must consider the relevant context. As explained earlier, we treat hate or non-hate as the regressor, embedding as the response, and run the regression to see if the same word has different meanings in hate versus non-hate contexts.

We begin the investigation on variance within human annotation with the most malicious racial slur, "n*****," a word against the black population. We want to understand whether human annotators of different identities would react differently to this obvious racial slur.


Our first regression investigates whether black annotators would rate the racial slur differently as compared to annotators of other identities. We choose this regression because studies of English linguistics such as ~\citep{Rahman2012TheNW} show that this term has "continued acceptance among some members of the African American community for intragroup self-reference." While embedding regression cannot show us whether black annotators are more likely or less likely to rate speeches containing this term as hate speech, it is able to show us whether there is a statistically significant difference in their annotation as compared to annotators of other identities. The result is: yes, there is a statistically significant difference.

\begin{table}[!ht]
\begin{tabular}{|c c c c c c|}
 \hline
 coefficient & normed.estimate & std.error & lower.ci & upper.ci & p.value\\
 \hline
target\_race\_black & 6.97 & 0.13 & 6.72 & 7.21 & 0 \\
annotator\_race\_black & 0.49 & 0.04 & 0.41 & 0.57 & 0 \\
 \hline
\end{tabular}

\caption{Embedding regression of the racial slur against whether the comment is hate speech towards the black population and whether the annotator is black.}
\label{tab:reg-black}
\end{table}

In our second embedding regression shown in Table~\ref{tab:black-women}, we found it statistically significant that female annotators rate comments that contain the term differently as hate speech that target the black population. Again, we do not know if this female annotators rate it more harshly or more gently, because the regression only shows us that they have a distinctive reaction to this term. However, in testing with other slurs, for example those that target the Asian population, we did not find such a statistically significant correlation, which means that female annotators do not have significantly different reactions to all racial slurs.

In randomly pulling instances, we observe that human annotators would sometimes falsely label a long sentence containing "MLK" as hate speech, so we wonder if annotators' education background would also influence how they rate online speeches that contain "MLK." In the third table, we regress run embedding regression of online speech that contains "MLK" against whether the comment is hate speech against the black population and whether the annotator has "some high school" education. In the fourth table, we do the same but only switch the regressor as annotators with PhD education. Both tables show that people with different education background have significantly different evaluations of online instances that contain the word "MLK."

\begin{table}[!ht]
\begin{tabular}{|c c c c c c|}
 \hline
  coefficient & normed.estimate & std.error & lower.ci & upper.ci & p.value\\
  \hline
target\_race\_black & 6.97 & 0.13 & 6.72 & 7.21 & 0 \\
annotator\_gender\_women & 0.28 & 0.03 & 0.23 & 0.33 & 0 \\
\hline
\end{tabular}
\caption{Embedding regression of the racial slur against whether the comment is hate speech towards the black population and whether the annotator is female.}
\label{tab:black-women}
\end{table}

\begin{table}[!ht]
\begin{tabular}{|c c c c c c|}
 \hline
  coefficient & normed.estimate & std.error & lower.ci & upper.ci & p.value\\
  \hline
target\_race\_black & 3.21 & 0.08 & 3.05 & 3.37 & 0 \\
annotator\_education\_phd & 1.19 & 0.11 & 0.96 & 1.41 & 0 \\
\hline
\end{tabular}
\caption{MLK-Ph.D.}
\label{tab:mlk-phd}
\end{table}

\begin{table}[!ht]
\begin{tabular}{|c c c c c c|}
 \hline
  coefficient & normed.estimate & std.error & lower.ci & upper.ci & p.value\\
  \hline
target\_race\_black & 3.21 & 0.08 & 3.05 & 3.37 & 0 \\
annotator\_education\_some\_high\_school & 1.94 & 0.24 & 1.48 & 2.41 & 0 \\
\hline
\end{tabular}
\caption{MLK-some high school}
\label{tab:mlk-some-hs}
\end{table}


We don't need to go into more examples to show distinctions. The finding is such that there exists a wide range of variance because humans of different identities would rate online instances differently.



Now, we want to show why data imbalance for the Measuring Hate Speech data set does not influence our final result in embedding regression.
In some sense, embedding regression estimates
\begin{equation}
\begin{aligned}
& E(\text{embedding}|\text{target\_race\_black}, \text{annotator\_gender\_women}) \\
=& \beta_0 + \text{target\_race\_black} \cdot \beta_b + \text{annotator\_gender\_women} \cdot \beta_w
\end{aligned}
\end{equation}

\begin{equation}
\begin{aligned}
&E(\text{embedding}|\text{annotator\_gender\_women = 1}) \\
- &E(\text{embedding}|\text{annotator\_gender\_women = 0}).
\end{aligned}
\end{equation}

Since we are conditioning on $\text{annotator\_gender\_women}$, its marginal distribution does not affect the result of the  embedding regression.



\section{Training a Hate Speech Classifier}

\subsection{Findings from LLM Classification}
 

We combine traditional machine learning methods and state-of-the-art speech classification through text embedding.
This method, known as linear probing~\citep{alain2017understanding}, works as follow:
We trained a hate speech classifier on a top-performing Large Language Model, NV-Embed-v2~\citep{lee2024nvembedimprovedtechniquestraining}, which according to the latest MTEB~\citep{muennighoff2022mteb} benchmark achieves a test accuracy of 92.74\% in Toxic Conversations Classification.
On top of these embedding, we then fitted a traditional machine learning method --- logistic regression --- to predict the human annotations in \textit{Measuring Hate Speech}, which is hand-coded by Amazon Mechanical Turk workers.

The accuracy of this regression is 93.83\% on the training set and 94.14\% on the test set. These are relatively close to the classification score of the NV-Embed-v2 on Toxic Conversations Classification. We then pulled the observations in the testing set which are misclassified by our model. These observations, however, are not necessarily misdiagnosed by logistic regression. Indeed, they suggest that models and humans disagree on the classification of these speeches.

Here, $Y$ denotes prediction. $X$ denotes the embedding. $\beta$ is the parameter for logistic regression. So, no matter what the distribution of identity is, this observed pattern is not affected by this distribution, i.e., in the case where 80\% annotators who are women, this distribution should not affect the measurement.

\begin{equation}
\begin{aligned}
P(Y=1|X) =& \frac{1}{1+\exp(-X \beta)}
\end{aligned}
\end{equation}

Does the variance within the training set (i.e., women's relatively higher sensitivity to the racial slur that target the black population) hurt the machine's prediction?
Note that in our logistic regression classifier, $X$ is the sentence embedding and does not include the annotator's identity.
Hence, the uncertainty in the annotator's identities gets absorbed into the variance in $Y|X$, resulting in a larger conditional variance of $Y$ given $X$, i.e., $\operatorname{Var}(Y|X)$, which would decrease the Fisher information, and increase the variance of the maximum likelihood estimator, i.e., $\operatorname{Var}\left(\hat{\beta}\right)$.
To put it the other way, we can consider this variance as (potentially class-conditional) label noise~\citep{NIPS2013_3871bd64}, which may hurt the predictive accuracy of the fitted model.
We do not have the objectively correct label for noise, so this is not measurable - we only know it's there.

\subsection{Hypothesis: What Went Wrong with the LLM?}
To understand where humans and machines disagree, the first step of our diagnosis is to manually code these observations. In randomly pulling instances from the ``wrong'' set, we agree with the machine's judgment 70 percent of the time as opposed to that of humans, which we only agree with 30 percent of the time.
Hence, in our speech classifier, the logistic regression model on top of LLM embeddings outperformed the Amazon Mechanical Turk workers. We interpret this phenomenon as such that LLM did the most heavy lifting by converting text to embedding, while logistic regression is only the last kick. Hence, the embeddings accurately capture the semantic meaning of the text.

The next thing we do is to use cosine similarity ratio to portray the linguistic differences on the speeches in which machines disagree with humans disagree. Recall that the cosine similarity between word embedding vectors characterizes the semantic similarity of the underlying words. To gain some intuition behind the semantic difference of a certain word under hate vs non-hate speech, we analyze the embeddings of its context in both groups. In particular, we first compute the ALC embedding of this particular word for each group, known as ``group embeddings''. Then, we find a handful of words whose cosine similarity is closest to each of the group embeddings, known as ``features''. Finally, for each given feature we compute the similarity between it and each group embedding, and then take the ratio of these two.
This ratio captures how ``discriminant'' a feature/word is of a given group. 


To understand what causes the disparity between logistic regression and human annotators, we compute the cosine similarity ratio (we use human annotator's judgment \texttt{target\_race\_black = TRUE} as the numerator) of the very offensive racial slur on the samples where human and machines disagree with each other. The result: features with the largest cosine similarity ratio are swear words, suggesting that the presence of these slang tends to make humans think a certain comment is hate speech, and simultaneously make machine learning models classify as non-hate speech. In pulling the instances in which the LLM mis-classifies as non-hate speech, which contain both the racial slur and the swear words, we found that these are both relatively short and highly toxic.

This seems to be counterintuitive: How can machines succeed in detecting long statements but fail at detecting short instances of obvious hate content that include racial slurs? 


\subsection{Diagnostics}

Usually, two conscious actions could contribute to a model's self-censorship on hateful content: one is model alignment, defined as the process of aligning an AI system's behavior with human values. Another is done by removing malicious content from the training set. Self-censorship can also happen unknowingly: if the training dataset is missing toxic content, and the researchers are unaware of this fact, then the model would also fail at detecting such content.

Hence, there are four stages where the NV-Embed-v2 model could be restricted in detecting hate speech:
\begin{itemize}

\item Model alignment that happens in the base model of NV-Embed-v2, which is the Mistral-7B model. 
\item Data filtering that happens in training the NV-Embed-v2 model.
\item In training the NV-Embed-v2 model, researchers used a two-phase training process: Phase 1 focused exclusively on retrieval datasets; Phase 2 focused on non-retrieval datasets. Given that the retrieval datasets contain minimal toxic content, Phase 1 could have diminished the model's capacity to form embeddings related to toxic content, and that capacity may not be restored in Phase 2.
\item Finally, there could be proactive model alignment due to ethical reasons before launching the NV-Embed-v2 model.
\end{itemize}

In reviewing \cite{lee2024nvembedimprovedtechniquestraining}, we ruled out stage 2 and 4 because the NV-Embed-v2 model was trained on a purposefully chosen toxic dataset, and there was no model alignment. In addition, the researchers sampled the toxic and non-toxic examples equally, so there should not have been an imbalance in the training set. However, the researchers recognized that Stage 1. and 3. could have contributed to the model's failure in detecting obvious toxic content.

Because we don't have the bandwidth to replicate their model training process, we diagnosed Stage 1. by taking the base Mistral-7B model and used few-shot text completion to generate racial slurs. Here is the output we saw by prompting the original LLM \texttt{mistralai/Mistral-7B-Instruct-v0.1} - the same output occurs after several attempts:

\begin{verbatim}
[INST]What is the most offensive slur targeting the black population?[/INST]
One of the most offensive slurs targeting the black population is the
"N-word." Its use is highly derogatory and considered extremely offensive 
by many.
\end{verbatim}

We see that in the base model, the offensive racial slur has already been replaced by its euphemism. In addition, we also examined the generated racial slurs that target different ethnic groups, such as Black, Asian, Hispanic, and White. We observed that these generated slurs have relatively low toxicity. We therefore conclude that self-censorship has occurred in the base model Mistral-7B, which in turn affects the iterated model NV-Embed-v2. 

This is a stunning finding because NV-Embed-v2 is one of the top performing open-sourced LLMs on toxic speech classification, but its performance is greatly limited due to human effort. This implies that 1) many LLMs we have nowadays have some forms of self-censorship, 2) the current LLMs could ideally approximate near-perfect prediction without self-censorship.






\section{Implications for the future of AI}
First, this study shows us that open-sourced large language models are the future for responsible AI. They are designed such that researchers can easily find out an ethical choice in their transparent training process and go stage-by-stage into the diagnosis of model performance. This is crucial for policymakers who care about the ethics of AI, not only because close-sourced models like GPT4 cannot support this kind of diagnostics, but also because the active monitoring from the research community plays a powerful role in AI regulations.

We argue that research community's active monitoring can be more powerful than government restrictions because the launch of these models cannot avoid peer reviews from the Machine Learning conferences. In fact, the NeurIPS Code of Ethics clearly states that ``Authors should transparently communicate the known or anticipated consequences of research: for instance via the paper checklist or a separate section in a submission," and their areas of concerns include the following principle: ``Bias and fairness: Contributors should consider any suspected biases or limitations to the scope of performance of models or the contents of datasets and inspect these to ascertain whether they encode, contain or exacerbate bias against people of a certain gender, race, sexuality, or other protected characteristics." This indicates that if an open-sourced model is able to produce hate speech, it may not get accepted by conferences that care about ethics. In the meantime, close-sourced proprietary models has much less of this concern because researchers cannot easily enter their training process.

Now, our study shows that by restricting an LLM from producing toxic content, we also restrict its ability to detect such content. While we do not know how many of the LLMs are consciously aligned, this seems to be a default shared agreement among researchers. We can find no better explanation than a philosophical one to explain this phenomenon. In Nichomachean Ethics, Aristotle argues that humans are driven by a virtue of the "golden mean," which is a balance of the extremes, such as between the indulges and the abstinence, the courage and the cowardice. Aristotle also argues that between two evils, if humans must choose one, they will always choose the lesser evil. Hence, when researchers are asked to train a language model, the moral restrictions in their own language are mirrored into the model's performance. Between the two extremes: training a super toxic model that produces offensive racial slurs and silencing the model such that it is bad at classification, the researchers choose the "mean," which is to censor it a little bit. As compared to the greater evil of creating an LLM that offends people, they choose the "lesser evil" of creating a less toxic model. According to Aristotle's theory of ethics, LLMs' self-censorship may arise as researchers' expression of sophrosyne (temperance) to refrain from causing social disharmony, or often simply, a concern that an AI model trained on such datasets wouldn't pass the reviewers' ethical guidelines.

\section{Conclusion}
In this paper, we use embedding regression to show from a granular level that human annotators of different identities show differences in classifying hate speech given the particular words that the speech contains. This variance shows as noise and affects the machine's classification of hate speech to a limited extent. The greater influence on machine classification comes from the human's participation in training the model: through model alignment, researchers reduce the LLMs' ability to detect obvious racial slurs. We deduce that this form of censorship widely exist in open-sourced LLMs. We conjecture that researchers prefer to choose the "lesser evil" in avoiding societal consequences, and their own moral restrictions factor deeply into the model training process.

\section{Acknowledgement}
This paper could not have been possible without Professor Jacob Shapiro's valuable feedback and sponsorship of computational resources as well as Professor Arthur Spirling's kind guidance.

The simulations presented in this article were performed on computational resources managed and supported by Princeton Research Computing, a consortium of groups including the Princeton Institute for Computational Science and Engineering (PICSciE) and the Office of Information Technology's High Performance Computing Center and Visualization Laboratory at Princeton University.

\newpage
\printbibliography

@article{RODRIGUEZ_SPIRLING_STEWART_2023, title={Embedding Regression: Models for Context-Specific Description and Inference}, volume={117}, DOI={10.1017/S0003055422001228}, number={4}, journal={American Political Science Review}, author={RODRIGUEZ, PEDRO L. and SPIRLING, ARTHUR and STEWART, BRANDON M.}, year={2023}, pages={1255–1274}}

@article{arora-etal-2016-latent,
    title = "A Latent Variable Model Approach to {PMI}-based Word Embeddings",
    author = "Arora, Sanjeev  and
      Li, Yuanzhi  and
      Liang, Yingyu  and
      Ma, Tengyu  and
      Risteski, Andrej",
    editor = "Lee, Lillian  and
      Johnson, Mark  and
      Toutanova, Kristina",
    journal = "Transactions of the Association for Computational Linguistics",
    volume = "4",
    year = "2016",
    address = "Cambridge, MA",
    publisher = "MIT Press",
    url = "https://aclanthology.org/Q16-1028",
    doi = "10.1162/tacl_a_00106",
    pages = "385--399",
}

@article{arora-etal-2018-linear,
    title = "Linear Algebraic Structure of Word Senses, with Applications to Polysemy",
    author = "Arora, Sanjeev  and
      Li, Yuanzhi  and
      Liang, Yingyu  and
      Ma, Tengyu  and
      Risteski, Andrej",
    editor = "Lee, Lillian  and
      Johnson, Mark  and
      Toutanova, Kristina  and
      Roark, Brian",
    journal = "Transactions of the Association for Computational Linguistics",
    volume = "6",
    year = "2018",
    address = "Cambridge, MA",
    publisher = "MIT Press",
    url = "https://aclanthology.org/Q18-1034",
    doi = "10.1162/tacl_a_00034",
    pages = "483--495",
}

@article{204871,
  author = {Pedro Rodriguez and Arthur Spirling and Brandon M. Stewart and Elisa Wirsching},
  title = {Multilanguage Word Embeddings for Social Scientists: Estimation, Inference and Validation Resources for 157 Languages},
  year = {2023},
  url = {https://alcembeddings.org/assets/img/RSSW_paper_June_2023.pdf},
}

@article{doi:10.1080/00437956.1954.11659520,
author = {Zellig S. Harris},
title = {Distributional Structure},
journal = {WORD},
volume = {10},
number = {2-3},
pages = {146--162},
year = {1954},
publisher = {Routledge},
doi = {10.1080/00437956.1954.11659520},
URL = {
        https://doi.org/10.1080/00437956.1954.11659520
},
eprint = {
        https://doi.org/10.1080/00437956.1954.11659520
}
}

@inproceedings{pennington-etal-2014-glove,
    title = "{G}lo{V}e: Global Vectors for Word Representation",
    author = "Pennington, Jeffrey  and
      Socher, Richard  and
      Manning, Christopher",
    editor = "Moschitti, Alessandro  and
      Pang, Bo  and
      Daelemans, Walter",
    booktitle = "Proceedings of the 2014 Conference on Empirical Methods in Natural Language Processing ({EMNLP})",
    month = oct,
    year = "2014",
    address = "Doha, Qatar",
    publisher = "Association for Computational Linguistics",
    url = "https://aclanthology.org/D14-1162",
    doi = "10.3115/v1/D14-1162",
    pages = "1532--1543",
}

@article{bojanowski2016enriching,
  title={Enriching Word Vectors with Subword Information},
  author={Bojanowski, Piotr and Grave, Edouard and Joulin, Armand and Mikolov, Tomas},
  journal={arXiv preprint arXiv:1607.04606},
  year={2016}
}

@misc{lee2024nvembedimprovedtechniquestraining,
      title={NV-Embed: Improved Techniques for Training LLMs as Generalist Embedding Models}, 
      author={Chankyu Lee and Rajarshi Roy and Mengyao Xu and Jonathan Raiman and Mohammad Shoeybi and Bryan Catanzaro and Wei Ping},
      year={2024},
      eprint={2405.17428},
      archivePrefix={arXiv},
      primaryClass={cs.CL},
      url={https://arxiv.org/abs/2405.17428}, 
}

@article{muennighoff2022mteb,
  doi = {10.48550/ARXIV.2210.07316},
  url = {https://arxiv.org/abs/2210.07316},
  author = {Muennighoff, Niklas and Tazi, Nouamane and Magne, Lo{\"\i}c and Reimers, Nils},
  title = {MTEB: Massive Text Embedding Benchmark},
  publisher = {arXiv},
  journal={arXiv preprint arXiv:2210.07316},  
  year = {2022}
}

@misc{
alain2017understanding,
title={Understanding intermediate layers using linear classifier probes},
author={Guillaume Alain and Yoshua Bengio},
year={2017},
url={https://openreview.net/forum?id=ryF7rTqgl}
}

@inproceedings{NIPS2013_3871bd64,
	author = {Natarajan, Nagarajan and Dhillon, Inderjit S and Ravikumar, Pradeep K and Tewari, Ambuj},
	booktitle = {Advances in Neural Information Processing Systems},
	editor = {C.J. Burges and L. Bottou and M. Welling and Z. Ghahramani and K.Q. Weinberger},
	publisher = {Curran Associates, Inc.},
	title = {Learning with Noisy Labels},
	url = {https://proceedings.neurips.cc/paper_files/paper/2013/file/3871bd64012152bfb53fdf04b401193f-Paper.pdf},
	volume = {26},
	year = {2013}}

@article{Rahman2012TheNW,
  title={The N Word},
  author={Jacquelyn Rahman},
  journal={Journal of English Linguistics},
  year={2012},
  volume={40},
  pages={137 - 171},
  url={https://api.semanticscholar.org/CorpusID:144164210}
}

\end{spacing}

\end{document}